\documentclass[letterpaper]{article}
   \usepackage{aaai24} 
   \usepackage{times} 
   \usepackage{helvet} 
   \usepackage{courier} 
   \usepackage[hyphens]{url} 
   \usepackage{graphicx} 
   \usepackage{graphicx}  
   \usepackage{natbib}  
   \usepackage{caption}  
   \usepackage{amsfonts}
   \usepackage{algorithm}
   \usepackage{algpseudocode}
\title{Clustered Policy Decision Ranking}
\author {
    Mark Levin\textsuperscript{\rm 1},
    Hana Chockler\textsuperscript{\rm 2}
}
\affiliations {
    \textsuperscript{\rm 1}University of Maryland, College Park, MD, US\\
    \textsuperscript{\rm 2}King's College, London, UK\\
    mlevin12@umd.edu, hana.chockler@kcl.ac.uk
}

\begin{document}

\maketitle

\begin{abstract}
Policies trained via reinforcement learning (RL) are often very complex even for simple tasks. In an episode with n time steps, a policy will make n decisions on actions to take, many of which may appear non-intuitive to the observer. Moreover, it is not clear which of these decisions directly contribute towards achieving the reward and how significant their contribution is. Given a trained policy, we propose a black-box method based on statistical covariance estimation that clusters the states of the environment and ranks each cluster according to the importance of decisions made in its states. We compare our measure against a previous statistical fault localization based ranking procedure.
\end{abstract}

\section{Introduction}

Reinforcement learning is a powerful 
method for training policies that
complete tasks in complex environments via sequential action selection \citep{Sutton2018}.
The policies produced are optimized to maximize 
the expected cumulative reward provided by the environment. While
reward maximization is clearly an important goal, 
this single measure may not reflect other objectives that an engineer or scientist may desire in training RL agents. Focusing solely on 
performance risks overlooking the demand for
models that are easier to analyse, 
predict and interpret~\citep{lewis2020deep}.
Our hypothesis is that many trained policies
are \emph{needlessly complex}, i.e., that there
exist alternative policies that
perform just as well or nearly as well
but that are significantly simpler.

The starting point for our definition of simplicity is
the assumption that there exists a way to make a \emph{simpler choice}
based on repeating the most recent action taken. We argue that this may be the case
for many environments in which RL is applied. That is, there may be states or clusters of states in which the most recent action can be repeated without a drastic drop in expected reward obtained. The tension between performance and simplicity is central to the field of explainable AI (XAI), and 
machine learning as a whole~\citep{gunning2019darpa}.


Ranking policy decisions according to their importance was introduced by~\citep{PCSK21}, who use spectrum-based fault localization (SBFL) techniques to approximate the contribution of decisions to reward attainment. This approach constitutes a statistical ranking of individual states.

We theorize however that a single state cannot have particularly notable importance, especially in more complex environments. As discussed in \citep{causal}, another exploration of policy simplification, previous work treats actions independently, whereas it is often the case that action sequences, or even non-contiguous combinations of actions, may synergize to have a large effect on reward acquisition \citep{Sutton1999b}. Extending SBFL-based and causality-based policy ranking into the multi-action domain may result in qualitatively distinct policy ranking results compared to single-action methods.

The key contribution of this paper is a novel method for clustering policy decisions according to their co-variable correlation with significant impact on the attainment of the goal, and subsequent ranking of those clusters. We argue that the clustered decisions grant insight into the operation of the policy, and that this clustered ranking method describes the 
importance of states more accurately than an individual statistical ranking. 
We evaluate our clustering method by using the clusters to simplify policies without compromising performance, hence addressing one of the main hurdles for wide adoption of deep RL: 
the high complexity of trained policies.

We use the same proxy measure for evaluating the quality of our ranking
as~\citep{PCSK21}: we construct new, simpler policies (``pruned policies'') that only use the top-ranked clusters,
without retraining, and compare the reward achieved by these policies to the original policy's reward. 

Our experiments with agents for MiniGrid~\citep{gym_minigrid} and environments from Atari Zoo~\citep{DBLP:journals/corr/BrockmanCPSSTZ16} including Bowling, Krull, and Hero
demonstrate that pruned policies can maintain high performance and also that performance monotonically approaches that of the original policy as more highly ranked decisions are included from the original policy.
As pruned policies are much easier to understand than the original policies,
we consider this a potentially useful method in the context of explainable RL.
As pruning a given policy does
not require re-training, the procedure is relatively lightweight.
Furthermore, the clustering of states itself provides an important insight into the relationships
of particular decisions to the performance of the policy overall.

The code for the experiments and full experimental results are available at 
\url{https://anonymous.4open.science/r/Clustered-Policy-Decision-Ranking}.

\section{Background}

\subsection{Reinforcement Learning}\label{sec:rl}


We use a standard reinforcement learning (RL) setup and assume that the reader is familiar with the basic concepts.
An \emph{environment} in RL is defined as a Markov decision process (MDP) with components $\left\{ S, A, P, R, \gamma\right\}$, where $S$ is the set of states $s$, $A$ is the set of actions $a$, $P$ is the transition function, $R$ is the reward function, and $\gamma$ is the discount factor. An agent seeks to learn a policy $\pi:S\rightarrow A$ that maximizes the total discounted reward. Starting from the initial state $s_0$ and given the policy $\pi$, the state-value function is the expected future discounted reward as follows:

\begin{equation}\label{eq:rlvalue}
V_{\pi}(s_0) = \mathbb{E}\left(\sum_{t=0}^{\infty}\gamma^tR(s_t,\pi(s_t),s_{t+1})\right).
\end{equation}

A policy $\pi: S \rightarrow A$ maps states to the actions taken in these states and may be  stochastic. We treat the policy as a black box, and hence make no further assumptions about $\pi$. 

\subsection{Mutation}

When we simplify a policy, we mutate some of its states such that the default action (repeating the previous action) is taken rather than the policy action. Given a state space S, a simplified policy has a ''mutated'' space $S_M \subset S$ and a mutually exclusive ''normal'' space $S_N \subset S$, such that any time a state is encountered in $S_N$ the policy action is taken and any time a state is encountered in $S_M$ the default action is taken.

\subsection{TF-IDF Vectorization}

TF-IDF, which stands for Term Frequency-Inverse Document Frequency, is a numerical statistic used in information retrieval and text mining to evaluate the importance of a word in a document relative to a collection of documents (corpus). The goal of TF-IDF is to quantify the significance of a term within a document by considering both its frequency within the document (TF) and its rarity across the entire corpus (IDF)~\citep{SJ-IDF}.

Given a corpus $C$, document $D$ and term $t$, we define $C(t)$ to be the number of documents containing $t$, and $D(t)$ to be the number of appearances of $t$ in $D$. The TF-IDF score of a term $t$ in document $D$ is calculated as follows.

\begin{equation}\label{eq:TF_orig}
\mbox{TF}(D,t) = D(t) / |D|
\end{equation}

\begin{equation}\label{eq:IDF_orig}
\mbox{IDF}(t) = \log(|C| / (C(t) + 1))
\end{equation}

\begin{equation}\label{eq:TF-IDF}
\mbox{TF-IDF}(D,t) = TF(D,t) * IDF(t)
\end{equation}

Each document $D$ is then represented as a vector with the entry corresponding to each term $t$ in the corpus being $\mbox{TF-IDF}(D,t)$. The resulting vectorized documents provide a way to identify the importance of terms within a document in the context of a larger corpus. It is commonly used in various natural language processing (NLP) tasks such as document retrieval, text classification, and information retrieval. Terms with higher TF-IDF scores are considered more significant to a particular document.

\subsection{Principal Component Analysis}

Principal Component Analysis (PCA)~\citep{H-PCA} is a dimensionality reduction technique widely used in statistics and machine learning. Its primary goal is to transform high-dimensional data into a new coordinate system, where the axes (principal components) are ranked by their importance in explaining the variance in the data. The first principal component captures the most significant variance, the second principal component (orthogonal to the first) captures the second most significant variance, and so on.

PCA starts by computing the covariance matrix of the standardized data. The covariance matrix summarizes the relationships between different features, indicating how they vary together. The next step involves finding the eigenvalues and corresponding eigenvectors of the covariance matrix. The eigenvectors represent the principal components and are listed in order of decreasing variance. By selecting a subset of these principal components with the highest variance, you can achieve dimensionality reduction while retaining the most important information in the data.

\section{Method}

\subsection{Random Sampling}

\begin{algorithm}
    \caption{Sample Trajectory}
    \textbf{Input}: $\mu$, $\tau$, env, policy, default\\
    \textbf{Output}: $S_M$ or $S_N$, averageReward
    \begin{algorithmic}[1]
        \State{Let $S_M$, $S_N$ = \{\}}
        \State{Let totalReward = 0}
        \For{\_ in range($\tau$)}
            \State{env.reset()}
            \While{not env.end()}
                \State{Let state = env.getEncodedState()}
                \If{state $\in S_M$}
                    \State{env.step(default)}
                \ElsIf{state $\in S_N$}
                    \State{env.step(policy)}
                \ElsIf{random() $<\mu$}
                    \State{$S_M$.add(state)}
                    \State{env.step(default)}
                \Else
                    \State{$S_N$.add(state)}
                    \State{env.step(policy)}
                \EndIf
                \State{totalReward += env.reward()}
            \EndWhile
        \EndFor
        \If{$\mu < 0.5$}
            \Return $S_M$, totalReward$/\tau$    
        \Else~
            \Return $S_N$, totalReward$/\tau$
        \EndIf
    \end{algorithmic}
    \label{alg:sample_trajectory}
\end{algorithm}

The naive approach to finding optimal clusters would be to determine a degree of desired simplification, such as by selecting the size of $S_N$, generate a suite of all potential sets $S_N$ of that size, and keep the set with the consequent trajectory of highest reward. For obvious reasons this is not practical. The quantity of all potential $S_N$ of size $|S_N|$ is $(^{\,\,\,|S|}_{|S_N|})$, which is $\Theta(|S|^{|S_N|})$. Instead we can randomly sample trajectories and extract aggregate data from them.
In order to generate a randomly sampled suite of $S_N$ sets, we define a mutation rate hyper-parameter $\mu$, a suite size hyper-parameter N, and a trials hyper-paremeter $\tau$, and perform N runs of Alg. \ref{alg:sample_trajectory}.
We may also employ an encoder that simplifies our state space to a set of abstract states, thereby decreasing the size of S and thus the search space for sets $S_N$. This encoder may for example include down-scaling or grey-scaling images that are input to the agent, or discretizing a continuous state space. The more the encoder simplifies the environment without losing important information, the better and faster this process performs. Particularly, in a continuous state space, a discretizing encoder is prerequisite for this process to work at all.

We define a ``+'' suite as a suite with a $\mu > 0.5$. For such suites, we record the sets $S_N$ corresponding to successful runs as defined by some condition unique to the environment and their average rewards.
We define a ``-'' suite as a suite with $\mu < 0.5$. For such suites, we record the sets $S_M$ corresponding to unsuccessful runs and their average rewards.
For any run, we generate a ``+'' suite and a ``-'' suite, each of size N, using mutation rates $\mu$ and $1-\mu$. The recorded sets in each form the two types of important sets, those that enable a large reward when all other states are mutated, and those that significantly decrease reward by being mutated.

\subsection{Vectorization}

We employ a modified version of the TF-IDF process to vectorize our randomly sampled suites. 
$C$ now refers to a randomly sampled suite, $D$ now refers to a cluster from that suite and $t$ to a state in the environment.
$R(D)$ refers to the reward achieved by cluster $D$ min-max normalized across suite $C$.
$T(C)$ refers to 1 for the ``-'' suite and 0 for the ``+'' suite.
Each state can only appear once in a cluster, so $D(t)$ is now 1 if $t \in D$ and 0 otherwise.

\subsubsection{TF}

We modify Eq.~\ref{eq:TF_orig} as follows.

\begin{equation}\label{eq:TD_mod}
\mbox{TF}(D,t) = D(t)(R(D)^2 - T(C))
\end{equation}

By squaring $R(D)$, we emphasize the differentiation between high scoring and low scoring trajectories. Additionally, by subtracting $T(C)$, states in small groups that enabled large reward will have large positive TF scores in that vector, and states in small groups that, when mutated, caused a large decrease in reward will have large negative TF scores in that vector. These are our two criteria for states being important, so by doing this, we ensure that states that fulfil both criteria will have greater variance than those that fulfil only one, and thus be more significant to Principal Component Analysis.

\subsubsection{IDF}

Typical IDF makes it so that terms that appear often across documents have very low scores. In fact, a term that is in every document would receive an IDF of 0. For our purposes these states should be down-weighted, but not to such a high degree. We define a downweighting hyper-parameter $\delta$ and modify Eq.~\ref{eq:IDF_orig} as follows.

\begin{equation}\label{eq:IDF_mod}
\mbox{IDF}(t_k) = (\log_\delta(C(t_k) + \delta))^{-1}.
\end{equation}

Smaller $\delta$ leads to greater downweighting. We leave  Eq.~\ref{eq:TF-IDF} unchanged. By vectorizing using TF-IDF scores in the same way as described above, we now have two collections of vectorized clusters where important states will be represented by large positive or negative scores and states that were important together will have such large scores together across vectors.

\subsection{Cluster Extraction}

We would like to extract small clusters that appear together often in these high scoring vectors. We can create three matrices: The ``-'' matrix has the vectorizations of the ``-'' suite for its columns, the ``+'' matrix has the vectorizations of the ``+'' suite for its columns, and the ``+-'' matrix has the vectorizations of both suites for its columns.

By applying PCA to each of our three matrices we can get the principal components, each a linear combination of our states. Here, we define two more hyper-parameters: $\sigma$ is the number of clusters we extract, and $\eta$ is the proportion of states in a component to cluster. For each of the first $\sigma$ priincipal components of highest order, we extract a cluster consisting of the $\eta * |S|$ states with corresponding coefficients of the greatest absolute value in that component. Thus for each matrix we get $\sigma$ clusters each with $\eta * |S|$ states. For each cluster, we then execute runs of the policy pruned to where all states not in that cluster are mutated and record the average reward achieved. The clusters are ranked in order of decreasing average reward.

\begin{figure}[htb]
\centering
\includegraphics[width=.8\columnwidth]{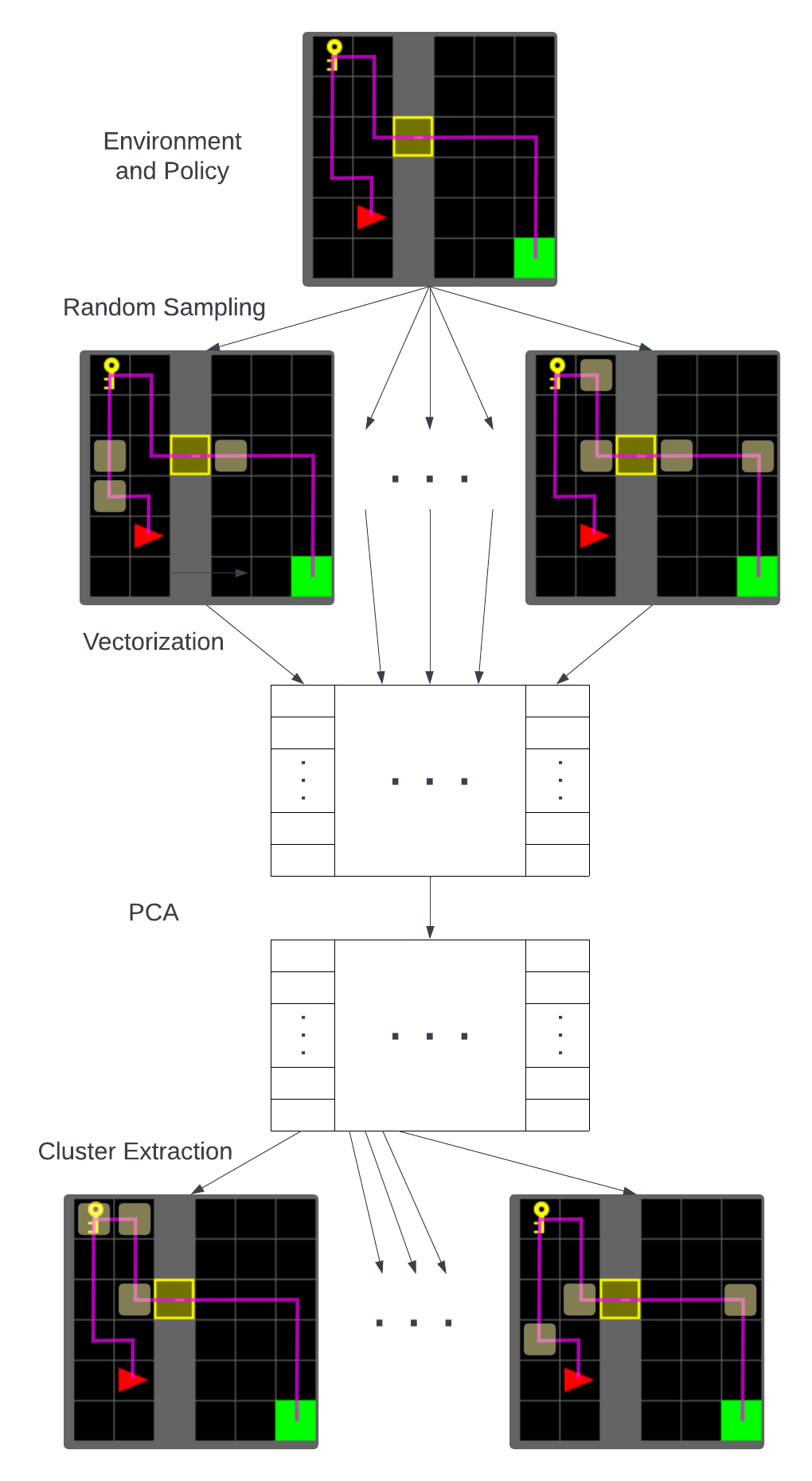}
\caption{A flowchart of the method, using Minigrid~\citep{gym_minigrid} as an example.}
\label{fig:flow}
\end{figure}

The clusters extracted from the ``+'' matrix form the ``cluster+'' ranking, those from the ``-'' matrix form the ``cluster-'' ranking, and those from the ``+-'' matrix form the ``cluster+-'' ranking,

A diagram of how the parts of the method fit together can be found in Fig~\ref{fig:flow}.

\section{Experimental Results}

\subsection{Experimental Setup}
We present the results of our experiments on a number of standard benchmarks.
The first benchmark is Minigrid~\citep{gym_minigrid}, a gridworld in which the agent operates with a cone of vision and can navigate many different grids, making it more complex than a standard gridworld. In each step the agent can turn or move forward.

The second set of experiments was performed on Atari games~\citep{DBLP:journals/corr/BrockmanCPSSTZ16}. We use policies that are trained using third-party code. No state abstraction is applied to the gridworld environments. For the Atari games, as is typically done, we crop the game’s border, grey-scale, down-sample to 18×14, and lower the precision of pixel intensities to make the enormous state space manageable. Note that these abstractions are not a contribution of ours, and were primarily chosen for their simplicity. For our main experiments, we use “repeat previous action” as the default action.
As a baseline, we compare our clustering method to the pre-existing processes to accomplish policy pruning, those being the SBFL, FreqVis, and Rand rankings explored in~\citep{PCSK21}.

We use the performance of pruned policies as a proxy for the quality of the ranking computed by our algorithm. In pruned policies, all but the top-ranked states are mutated. The pruned policies for SBFL, FreqVis, and Rand are designed as specified in ~\citep{PCSK21}. Beginning with all states mutated, an increasing fraction of states are ``restored'' (i.e. returned to the original) in each subsequent pruned policy, in order of decreasing rank.

Similarly, for each of our rankings ``cluster+'', ``cluster-'', and ``cluster+-'' we first run with all states mutated, and restore another cluster in each pruned policy in order of decreasing cluster rank. We then have two metrics by which to compare ranking processes. The first is what percent of the original reward can be accomplished given the proportion of the state space that is restored according to that process. The second is what percent of the original reward can be accomplished given what proportion of the actions taken in a trajectory are policy actions rather than default actions according to that process.

\subsection{Performance}

\begin{figure}[htb]
\centering
\includegraphics[width=.8\columnwidth]{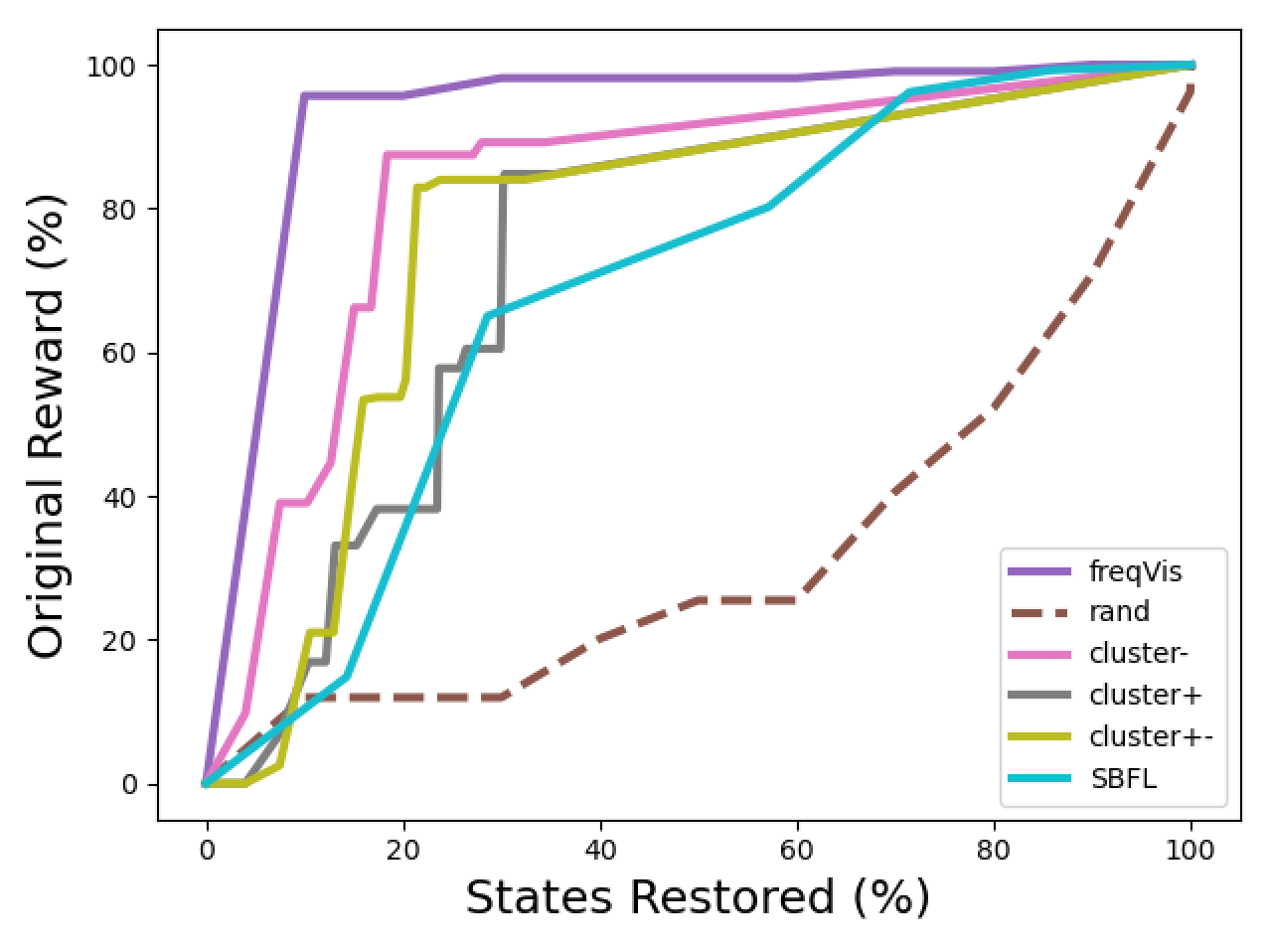}
\caption{Bowling Reward by States Restored}
\label{fig:bowling1}
\end{figure}

\begin{figure}[htb]
\centering
\includegraphics[width=.8\columnwidth]{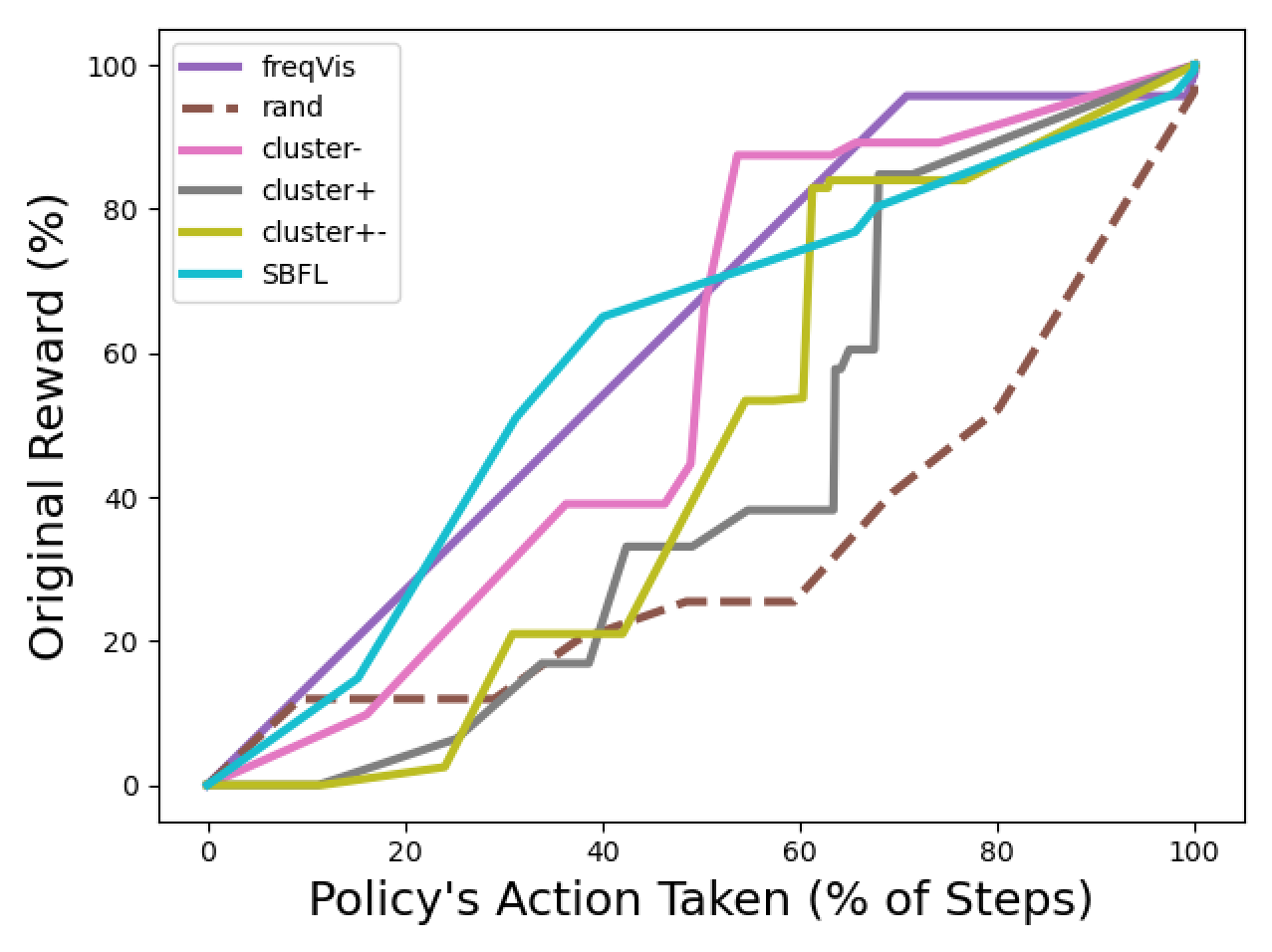}
\caption{Bowling Reward by Actions Restored}
\label{fig:bowling2}
\end{figure}

\begin{figure}[htb]
\centering
\includegraphics[width=.8\columnwidth]{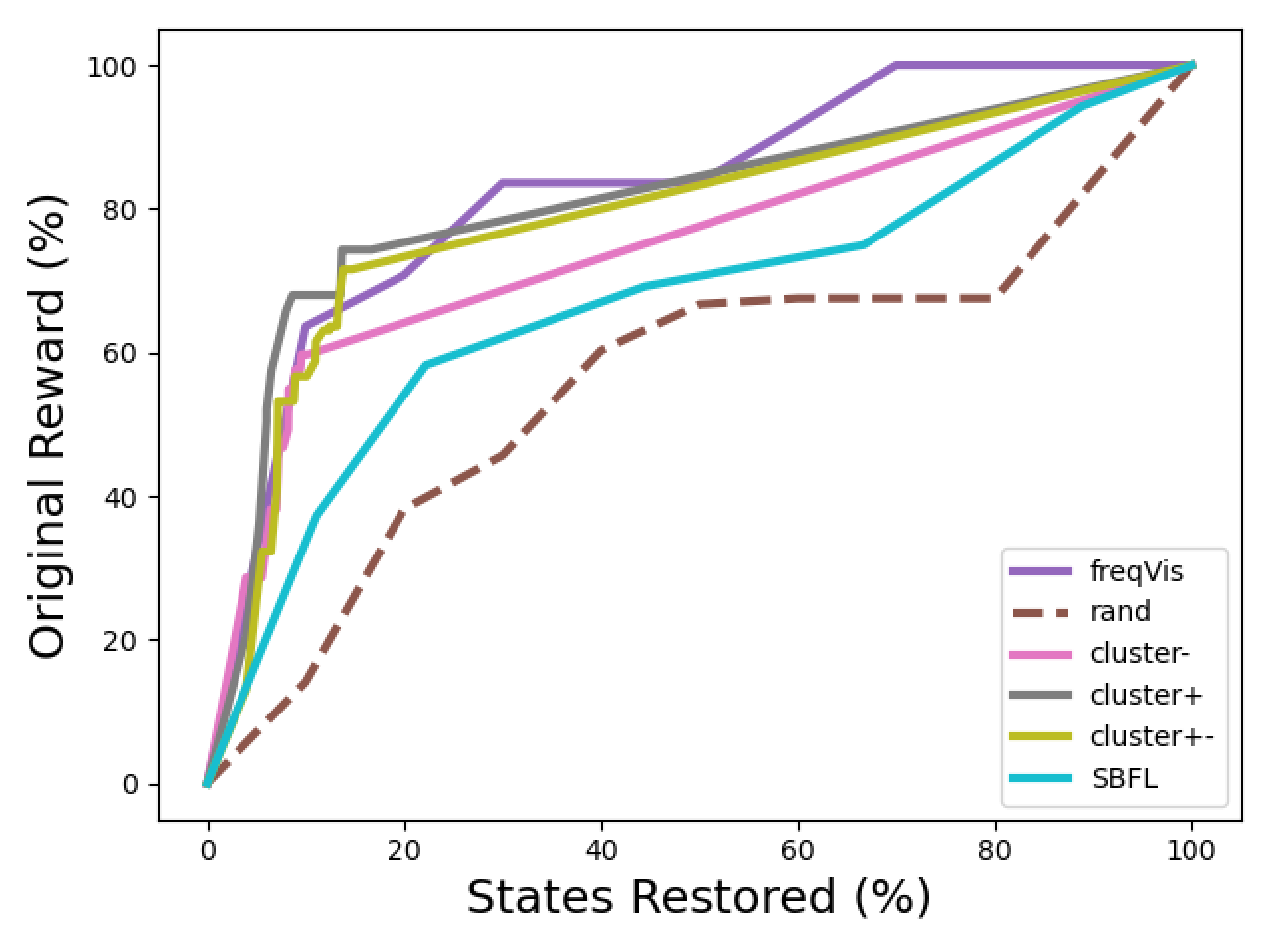}
\caption{Krull Reward by States Restored}
\label{fig:krull1}
\end{figure}

\begin{figure}[htb]
\centering
\includegraphics[width=.8\columnwidth]{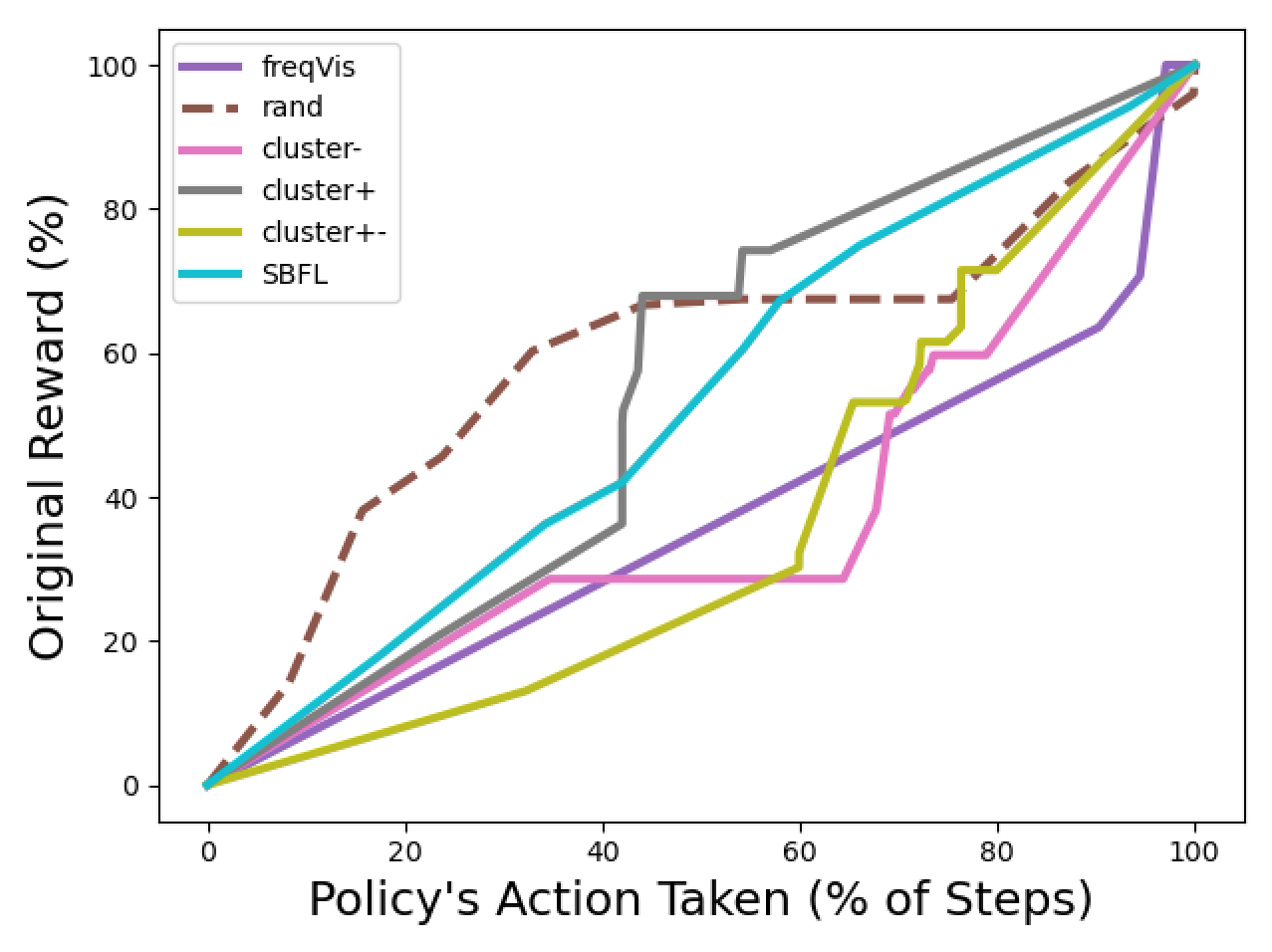}
\caption{Krull Reward by Actions Restored}
\label{fig:krull2}
\end{figure}

We find that in some environments one or more of our clustering methods outperforms SBFL, FreqVis, and Rand. There are many atari environments in which clustering outperforms SBFL but not FreqVis or vice-versa; the Bowling and Krull Atari games seemed best suited to the clustering process's strengths among the environments we explored.

Fig \ref{fig:bowling1} shows the comparison of methods by our first metric in the Bowling environment. All three clustering processes noticeably outperform SBFL, accomplishing most of the original reward with a significantly smaller portion of the state space restored. They do underperform FreqVis by this metric, however that is typically to be expected, as by definition FreqVis restores most of the actions by restoring only a small part of the state space. Our second metric which is shown in Fig \ref{fig:bowling2} is more suited to comparing a ranking's effectiveness against FreqVis. Here, the ``cluster-'' method mostly keeps pace with FreqVis and even reaches 80\% reward before it (though FreqVis achieves $95\%$ reward faster).  

Fig \ref{fig:krull1} shows the comparative performance by our first metric in the Krull environment. As in Bowling, all three clustering methods noticeably outperform SBFL, but ``cluster+'' and ``cluster+-'' seem on par with FreqVis even in the first metric. The addition of the second metric in Fig \ref{fig:krull2} reveals that FreqVis is not in fact well suited to the Krull environment, and ``cluster+'' outperforms all other methods. Based on these results, we extrapolate that Krull has a set of states that are not observed significantly more often than others (as Freqvis extracts) and do not individually have a significant effect on reward (as SBFL extracts), but collectively synergize to restore a significant proportion of the reward without requiring the restoration of larger parts of the state space.

\section{Conclusions and Discussion}

FreqVis, SBFL, and our clustering procedure each likely have their own place in policy simplification depending on the size, complexity, and characteristics of the environment.
We suggest using a portfolio platform that includes a number of techniques when trying to find an optimal pruning of a policy.
It is likely that our clustering method works significantly better when a feature based encoder is employed rather than simple abstractions 
such as grey-scaling or discretizing. This may be feasible given recent work in model-based reinforcement learning such as the development of the CLIP algorithm~\citep{r-clip} which learns an off-policy encoding of the state space without losing significant information and has the potential to expose pseudo-features.

As stated in~\citep{PCSK21}, in order to fully demonstrate the effectiveness of policy simplification as a method of explainable AI, a user study would need to be conducted. However, similarly to that previous work, that is outside the scope of this paper. In the absence of such a study, there exist theoretical justifications for policy simplification. One such justification is that by isolating the most impactful decisions that the policy makes, we can significantly reduce the number of places to look for potential issues with the policy. Alternatively, the simplified policy that is generated may also be useful for reducing the work of a more fine-toothed process to grant even more insight into the policy, allowing it to look at how the policiy makes only its most important decisions.

\newpage

\bibliography{refs}

\end{document}